\documentclass{article}

\usepackage{CJKutf8} 
\usepackage{longcat_style}
\usepackage{adjustbox}
\usepackage[utf8]{inputenc} %
\usepackage[T1]{fontenc}    %
\usepackage{newunicodechar}
\usepackage{hyperref}       %
\usepackage{xcolor}
\usepackage[normalem]{ulem} %
\hypersetup{
    colorlinks=true,      %
    linkcolor=blue,      %
    urlcolor=blue,       %
    citecolor=blue,      %
    linkbordercolor=blue, %
    urlbordercolor=blue,
    citebordercolor=blue,
    pdfborderstyle={/S/U/W 1}, %
}

\usepackage{float}
\usepackage{url}            %
\usepackage{booktabs}       %
\usepackage{amsfonts}       %
\usepackage{nicefrac}       %
\usepackage{microtype}      %
\usepackage{lipsum}		%
\usepackage{graphicx}
\usepackage{natbib}
\usepackage{doi}
\usepackage{amsmath}
\usepackage{amssymb} %
\usepackage{xspace}
\usepackage{enumitem}
\usepackage{multirow}
\usepackage{subcaption} 
\usepackage{makecell}
\usepackage{hyperref, cleveref}
\usepackage{pifont}
\usepackage[inkscapelatex=false]{svg}

\setlist[itemize]{leftmargin=*}
\setlist[enumerate]{leftmargin=*}
\setlist[description]{leftmargin=*}

\definecolor{midnightgreen}{rgb}{0.0, 0.29, 0.33}

\title{UniHetero: Could Generation Enhance Understanding for Vision-Language-Model at Large Data Scale?}

\author{
  Fengjiao Chen\thanks{Most of the work has been done before 2025.3. Please feel free to contact chenfengjiao02@meituan.com for detailed discussion or cooperation.} \quad
  Minhao Jing \quad
  Weitao Lu \quad
  Yan Feng \quad
  Xiaoyu Li \quad
  Xuezhi Cao \\
  Meituan, Beijing, China. }

\clearpage

\renewcommand{\shorttitle}{UniHetero: Generation Enhance Understanding At Scale}

\begin{document}
\maketitle

\begin{abstract}
\textbf{Takeaway 1: Generation can improve understanding, but Only if you generate Semantics, Not Pixels
}
A common assumption in unified vision-language models is that adding generation will naturally strengthen understanding. However, this is not always true at scale.
At 200M+ pretraining samples, generation helps understanding only when it operates at the semantic level, i.e. when the model learns to autoregress high-level visual representations inside the LLM. Once pixel-level objectives (e.g., diffusion losses) directly interfere with the LLM, understanding performance often degrades.



\textbf{Takeaway 2: Generation reveals a superior Data Scaling trend and higher Data Utilization.}
Unified generation-understanding demonstrates a superior scaling trend compared to understanding alone, revealing a more effective way to learn vision-only knowledge directive from vision modality rather than captioning to text.

\textbf{Takeaway 3: Autoregression on Input Embedding is effective to capture visual details.}
Compared to the commonly-used vision encoder, make visual autoregression on input embedding shows less cumulative error and is modality independent, which can be extend to all modalities. The learned semantic representations capture visual information such as objects, locations, shapes, and colors; further enable pixel-level image generation.


\end{abstract}

\begin{figure}[h!]
    \centering
    \includegraphics[width=0.9\linewidth]{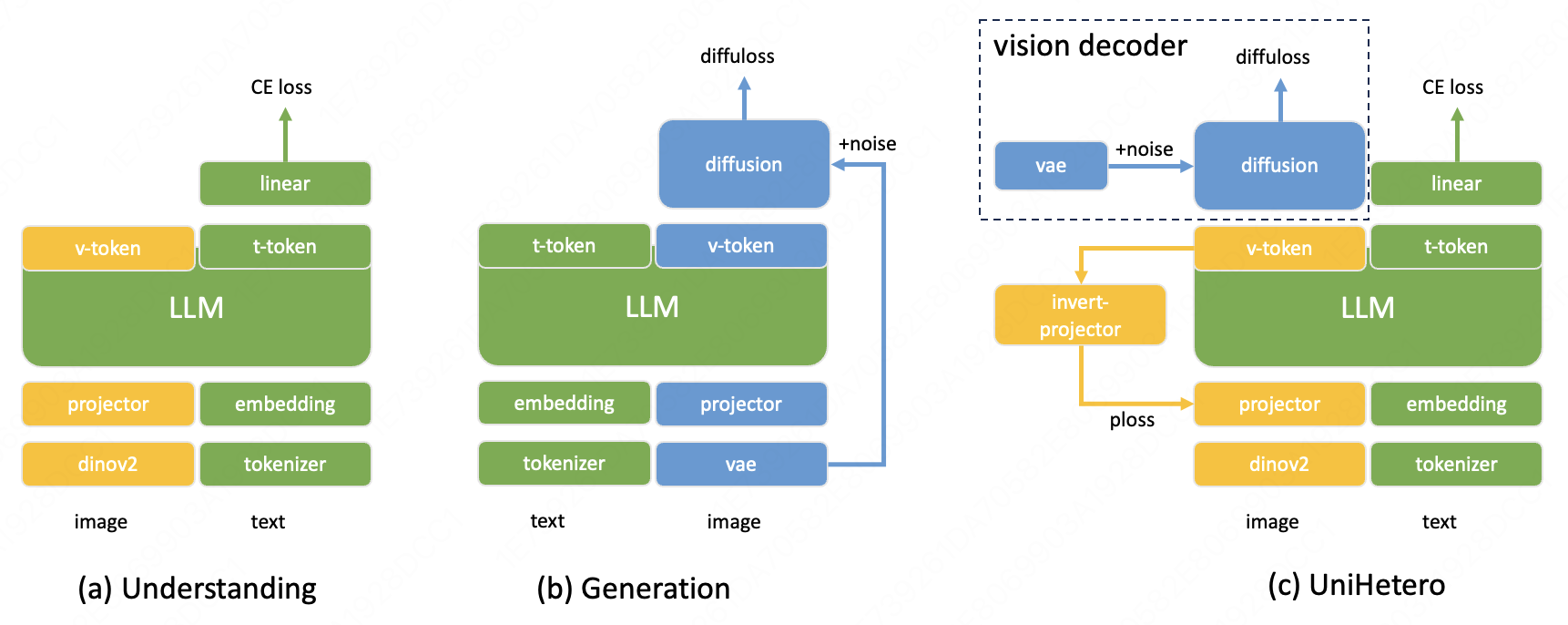}
    \caption{The unified vision language model structure UniHetero, which enables autoregression on semantic representation in LLM layer and transforms the pixel-level representation in modality related decoder structure.}
    \label{fig:model_structure}
\end{figure}

\clearpage
\tableofcontents
\clearpage

\section{Introduction}

With the rapid advancement of multimodal large models in visual understanding and visual generation, unified vision models have emerged as a promising trend for enhancing intelligence levels. 

Why are unified models considered favorable? From a model architecture perspective, visual generation structures are increasingly converging with visual understanding structures, with the core architecture predominantly being Transformer-based. Regarding the introduction of conditioning mechanisms, where differences are more pronounced, visual generation models are progressively simplifying, shifting from architecture-driven to data-driven approaches, particularly those capable of supporting multi-image or mixed image-text references. More importantly, from a data standpoint, the integration of tasks leads to an increase in the scale of utilizable data and improves data utilization efficiency under the same data scale.

In unified vision models, by integrating understanding tasks and generation tasks, it is "expected" that generation and understanding mutually reinforce each other, demonstrating a broader range of task formats and higher task performance. The term "expected" is used because the mutual enhancement between understanding tasks and generation tasks has not yet been fully validated, especially regarding the enhancement of understanding through generation.

Existing studies have validated the positive conclusion of "generation enhances understanding" in small-scale experimental settings ~\citep{tong2024metamorph,wang2024ross}. When the data scale is small, the enhancement brought by the data itself may have a greater impact compared to the task format.
However, in large-scale production scenarios (with training data exceeding 100M), the effect of generation tasks on understanding tasks is often negative~\citep{chameleonteam2025chameleon,wang2024emu3}. Several works~\citep{wu2024janus,ma2025unitok} discuss one of the reasons for the difficulty in integrating the two tasks: visual encoding differences. Understanding tasks require high-level semantic representations, while generation tasks require low-level texture representations. The disparity between these representations makes it challenging for existing visual encodings to accommodate both simultaneously. Improvements~\citep{ma2025unitok} attempt to fuse these two representations into a unified visual encoder, which shifts the difficulty from the LLM to the encoder which is challenging to solve as well.

To determine whether generation tasks can enhance understanding tasks, it is first essential to validate this in large-scale settings (>100M). 
In this work, we propose a concise autoregressive model with continuous representation and conduct a series of ablation experiments that verifies "generation enhances understanding" over 200M samples with better data-scale manner.
Main contributions
\begin{itemize}
    \item Demonstrates that generation enhances understanding on over 200M samples by relying on visual semantic representations rather than pixel-level representations.
    \item Introduces UniHetero, a concise autoregressive model with continuous representations that leverages existing modules to highlight the potential for improved data utilization efficiency in mainstream VLMs.
    \item Shows that performing autoregression directly on input embeddings is an effective strategy for modeling fine-grained visual details.
\end{itemize}

\section{Related Work}

\subsection{Unified Vision Language Models}
Vision-language large models are moving toward the unification of visual understanding and visual generation tasks. Existing unified vision models can be structurally categorized into end-to-end architectures and concatenated architectures. In end-to-end architectures, both understanding and generation tasks use the same visual encoder to feed visual information into the LLM backbone. The LLM’s parameters adapt to emphasize different tasks depending on the data, thereby achieving comprehensive task fusion~\citep{chameleonteam2025chameleon,wang2024emu3,tong2024metamorph,ge2025seedx}. In concatenated architectures, distinct visual encoders are used for understanding and generation tasks to feed visual information into the LLM backbone. The LLM’s parameters are explicitly partitioned by task, which prevents performance deterioration on individual tasks, but this design comes at the expense of limiting the possibility of task fusion~\citep{wu2024janus,deng2025bagel}. Additionally, there are unified models optimized specifically for generation tasks, aiming to enhance generation through understanding without preserving the performance of understanding tasks~\citep{xiao2024omnigen,tian2025unigen, zhou2024transfusion}. 

\subsection{Understanding Enhancement Challenge}
\label{sec:encoder_work}
Existing work has validated that understanding enhances generation~\citep{tian2025unigen,xiao2024omnigen}, but there is a lack of effective verification for generation enhancing understanding, especially under large-scale data pretraining. MetaMorph~\citep{tong2024metamorph}, with a sample size of 1M and training only during the SFT stage, demonstrated that as generation data increases, the metrics for understanding tasks show an upward trend. ROSS~\citep{wang2024ross}, pretrained on less than 1M samples, concluded through ablation studies that generation data enhances understanding tasks. When the data scale is small, the impact of the data itself on improvement is greater than the task format. However, in large-scale production scenarios (training data volume greater than 100M), the effect of generation tasks on understanding tasks is mostly negative~\citep{chameleonteam2025chameleon,wang2024emu3}. Prior studies~\citep{wu2024janus,ma2025unitok} point out that a key challenge in unifying these two tasks lies in their differing visual encodings: understanding tasks depend on high-level semantic features, whereas generation tasks require low-level texture details. This mismatch in required representations makes it difficult for current visual encoders to effectively support both types of tasks at once. UniTok~\citep{ma2025unitok} attempted to fuse these two representations by enhancing representational capabilities through multi-level codebooks, validating that generation can enhance understanding under discrete encoding with a sample size of 30M. 

Leading visual models in understanding primarily use continuous encoding, whereas unified visual encoders mainly rely on discrete encoding. By distilling existing continuous encoders to obtain semantic representations, they can approximate but hardly surpass them. At the same time, unifying visual representations is inherently challenging, effectively shifting the task fusion problem onto the visual encoder. To further explore the improvement of understanding effects through the enhancement of generation tasks based on existing visual understanding models, this work utilizes existing continuous encoders to construct a concise model structure, validating a better data-scale trend under large-scale pretraining.

\section{Methodology}
\subsection{Model Structure}
\begin{figure}[t]
    \centering
    \includegraphics[width=0.6\linewidth]{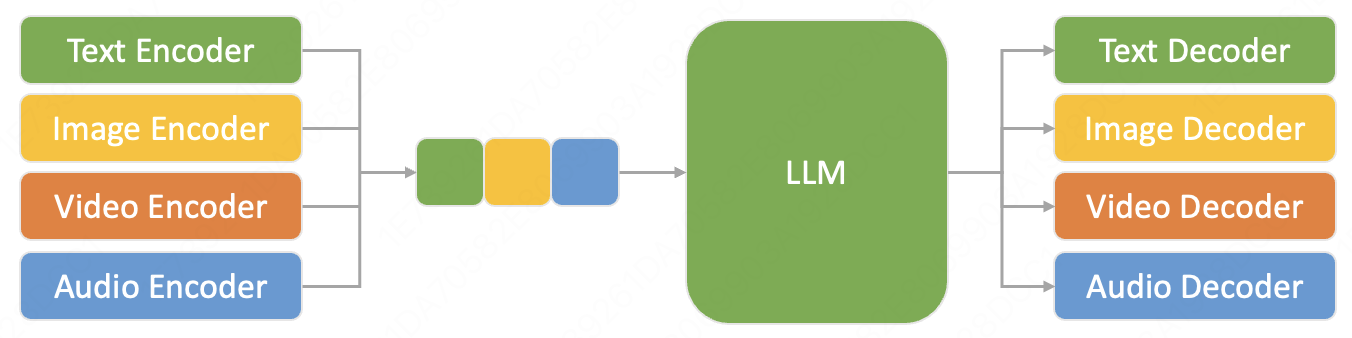}
    \caption{The ideal structure of multimodal large language model.}
    \label{fig:mllm_overview}
\end{figure}

To investigate the impact of generation tasks on understanding tasks, we propose integrating generation tasks into a visual understanding model while maintaining a concise architecture. This approach aims to preserve visual understanding capabilities while obtaining a multimodal autoregressive model. To mitigate conflicts among representations across different dimensions (in Section.\ref{sec:encoder_work}), we intend to fuse multimodal semantic representations inside the LLM and perform modality-specific generation through dedicated "decoders" for each modality, as shown in Figure.\ref{fig:mllm_overview}.

Based on the aforementioned considerations, we adopt the model architecture illustrated in Figure.\ref{fig:model_structure}. To better preserve comprehension capabilities, unlike~\citep{chameleonteam2025chameleon,wang2024emu3}, we employ continuous encoding to represent visual information, where VAE~\citep{li2024mar} represents visual generative representations (noted as pixel feature), and DINOv2~\citep{oquab2024dinov2} represents visual understanding representations (noted as semantic feature). At the LLM layer, only DINOv2 is used alongside text to perform multimodal auto-regression in the semantic dimension. In the vision decoder component, the image tokens output by the LLM serve as conditional inputs to the diffusion structure, which completes the details, decodes the image VAE features, and ultimately yields pixel-level images.





Consider the visual token with semantic representation $x^{s} \in \mathbb{R}^{ds}$ and pixel-level representation $x^{p} \in \mathbb{R}^{dp}$, which denotes the ground-truth token to be predicted. The LLM takes the aligned vector $e \in \mathbb{R}^{D}$ from the visual projector and generates the last hidden states $z \in \mathbb{R}^{D}$ at that same position. For semantic autoregression, we have Equation.~\ref{eq:auto_sematic_x}, which is also used in Seed-X~\citep{ge2025seedx} and MetaMorph~\citep{tong2024metamorph}).
\begin{equation}
\begin{split}
\mathcal{P}(x^{s}_{i} \mid x^{s}_{<i})  
&= \mathcal{P}(x^{s}_{i} \mid z_{i}) \, \mathcal{P}(z_{i} \mid x^{s}_{<i}) \\[4pt]
&= \mathcal{P}(x^{s}_{i} \mid z_{i}) \, \mathcal{P}(z_{i} \mid e_{<i}) \, \mathcal{P}(e_{<i} \mid x^{s}_{<i})
\end{split}
\label{eq:auto_sematic_x}
\end{equation}
Since the transformation from $e$ to $z$ reduces cumulative error, $z$ and $e$ have the same dimension, which is eaiser to learn. Therefore, we model the semantic autoregression on the input embedding $e$ of LLM in Equation.~\ref{eq:auto_sematic_e}. 
\begin{equation}
\begin{split}
\mathcal{P}(e_{i} \mid e_{<i}) 
&= \mathcal{P}(\hat{e}_{i} \mid z_{i}) \, \mathcal{P}(z_{i} \mid e_{<i})
\end{split}
\label{eq:auto_sematic_e}
\end{equation}
In this manner, all modalities can be treated within the same autoregressive framework, because $\mathcal{P}(z_{i} \mid e_{<i})$ does not depend on the modality. For $\mathcal{P}(\hat{e}_{i} \mid z_{i})$, it is determined solely by whether the variable is discrete or continuous, rather than by the modality itself.

Then the loss to fit semantic representation uses cosine similarity which shows better performances. More module selection experiments 
are shown in Section.~\ref{sec:module_selection}.
\begin{equation}
    \mathcal{L}_{\text{ploss}} = -cosine(e_{i}, \mathbb{E}_{P(\hat{e}_{i} \mid e_{<i})}\left[\hat{e}_{i} \right])
\end{equation}
And the loss to fit pixel-level representation is
\begin{equation}
    \mathcal{L}_{\text{diffuloss}} = \mathbb{E}_{\varepsilon, t} \left[ \left\| \varepsilon - \varepsilon_{\theta}(x^{p}_t \mid t, z) \right\|^2 \right]
    \label{eq:placeholder_label}
\end{equation}
Finally, we have the overall loss as
\begin{equation}
\mathcal{L} = \mathcal{L}_{\text{text}} + \alpha \mathcal{L}_{\text{ploss}} + \beta \mathcal{L}_{\text{diffuloss}}
\end{equation}
where $\alpha$ and $\beta$ are hyper-parameters to adjust the weight of loss.

\subsection{Training Details}
The training process consists of only two stages: pretraining and finetuning. The primary focus is on observing the effects of pretraining, while finetuning is employed solely to evaluate end-to-end performance metrics.

For pretraining, we employ an internal corpus of 80 million image–text aligned pairs tailored for image understanding, running for 3 epochs to yield 240 million training instances in total. Training batch size is 64 and the iteration amount is 37.5K. Throughout training, images are randomly positioned either before or after the text, with the latter arrangement treated as image generation data. Finetuning is then performed on a small internal dataset to initially elicit the model’s capabilities and support experimental evaluation.

We employ Llama2-7B~\citep{touvron2023llama2} as the LLM backbone and adopt the training configurations from Chameleon~\citep{chameleonteam2025chameleon} to facilitate modality fusion, including techniques such as qknorm and zloss. For image representations, semantic features are encoded using DINOv2-large~\citep{oquab2024dinov2}, while pixel-level features are encoded from MAR-KL16~\citep{li2024mar}. Both vision encoders are frozen during training to minimize confounding factors. The semantic features are integrated into the LLM via full-attention mechanisms, with masking probabilities sampled from a Gaussian Distribution (mean = 0.7, detailed in Section 3.3.2). As for vision decoder, we follow the similar way in MAR~\citep{li2024mar} for training and inference, where the pixel-level features are decoded using flow matching.

To balance the magnitude of losses across different modules, the weighting coefficients are set as $ \alpha = 10 $ and $ \beta = 10 $. 

\begin{figure*}
    \centering
    \includegraphics[width=\textwidth]{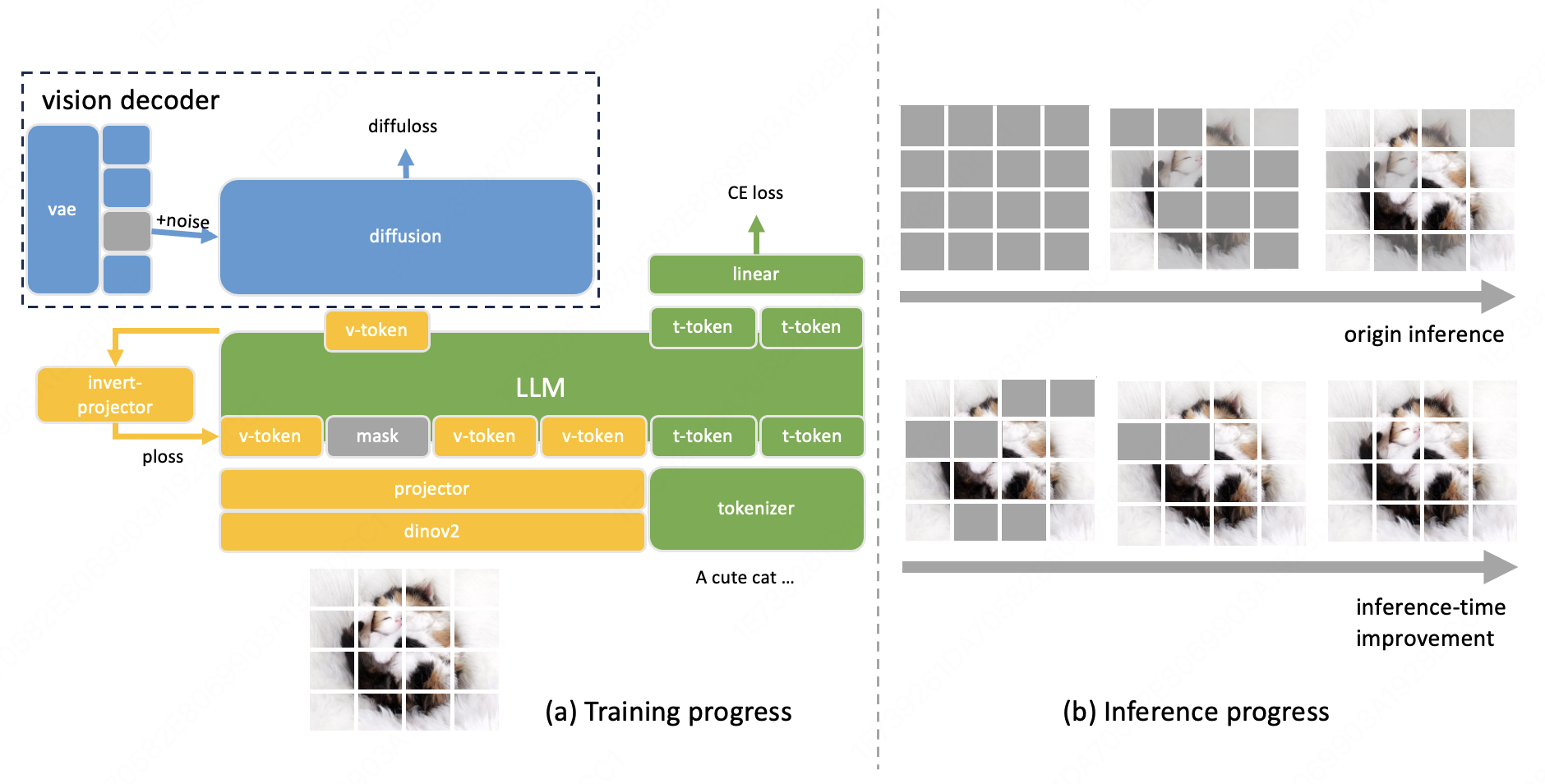}
    \caption{Illustration of training and inference progress in UniHetero. (a) shows the training progress with autoregression on input embedding rather than vision encoder and gaussian distribution for mask\_rate scheduler. (b) shows the inference progress where a inference-time improvement is proposed to refine the pixel-level generation without influence understanding ability.}
    \label{fig:train_and_infer}
\end{figure*}

\section{Experiment}
The experiments are designed to investigate: (1) whether generative modeling enhances understanding; (2) the semantic content encoded in visual representations; and (3) the capability of heterogeneous representations to synthesize images.

\subsection{Generation Enhances Understanding}
Under the UniHetero model, we analyze whether "generation enhances understanding" through ablation of the image generation loss. The experiment settings are shown in the Table.~\ref{tab:main_result}. The control group exp1 uses only the text generation loss, while the experimental groups exp2 and exp3, adds two types of image generation losses on top of the control group respectivelly. In terms of implementation, exp2 does not completely remove the diffusion loss but prevents gradients from back propagating to the LLM, thereby reducing the impact on image understanding while retaining image generation capabilities. 

To observe the data-scale trending quantitively, we use a linear regression to fit the performance in Equation~\ref{eq:linear_scale}.

\begin{equation}
\label{eq:linear_scale}
y = a n + b
\end{equation}

where $a$ indicates the data-scale trending, $n$ is the amount of data samples, $y$ is the evaluation metric.

\begin{table}[!ht]
  \centering
  \resizebox{0.8\columnwidth}{!}{
     \begin{tabular}{l|l|c|c|c|c|c|c}
     \toprule
        ID & Method & diffuloss & ploss & MMBench & MMBench($a$) & SeedBench & SeedBench($a$) \\
        \midrule
        exp1 & baseline   & 0 & 0 & 0.6357 & $-4\times10^{-4}$ & 0.6668 & $8\times10^{-4}$ \\
        exp2 & +ploss     & 0 & 1 & \textbf{0.6460} & \textbf{$66\times10^{-4}$} & \textbf{0.6696} & \textbf{$26\times10^{-4}$} \\
        exp3 & +diffuloss & 1 & 1 & 0.6168 & $62\times10^{-4}$ & 0.6456 & $7\times10^{-4}$ \\
        exp4 & +warmup    & 1 & 1 & 0.6271 & $30\times10^{-4}$ & 0.6585  & $7\times10^{-4}$\\
        \bottomrule
    \end{tabular}  
  }
  \caption{Ablation performences of UniHetero on various image generation strategies at more than 200M data-scale.}
  \label{tab:main_result}
\end{table}

\begin{figure*}[t]
    \centering
    \begin{subfigure}[b]{0.48\textwidth}
        \centering
        \includegraphics[width=\textwidth]{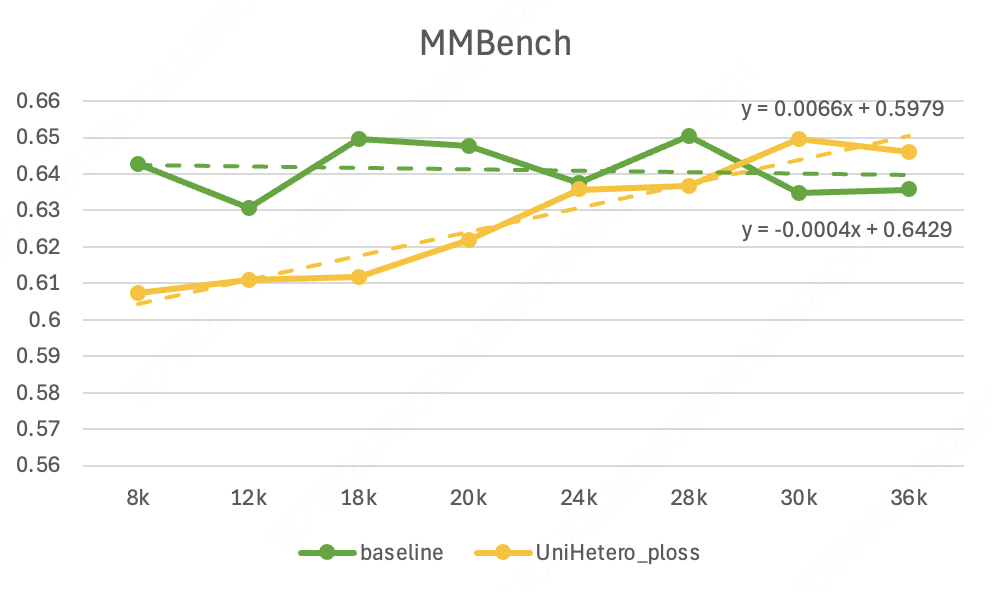}
    \end{subfigure}\hfill
    \begin{subfigure}[b]{0.48\textwidth}
        \centering
        \includegraphics[width=\textwidth]{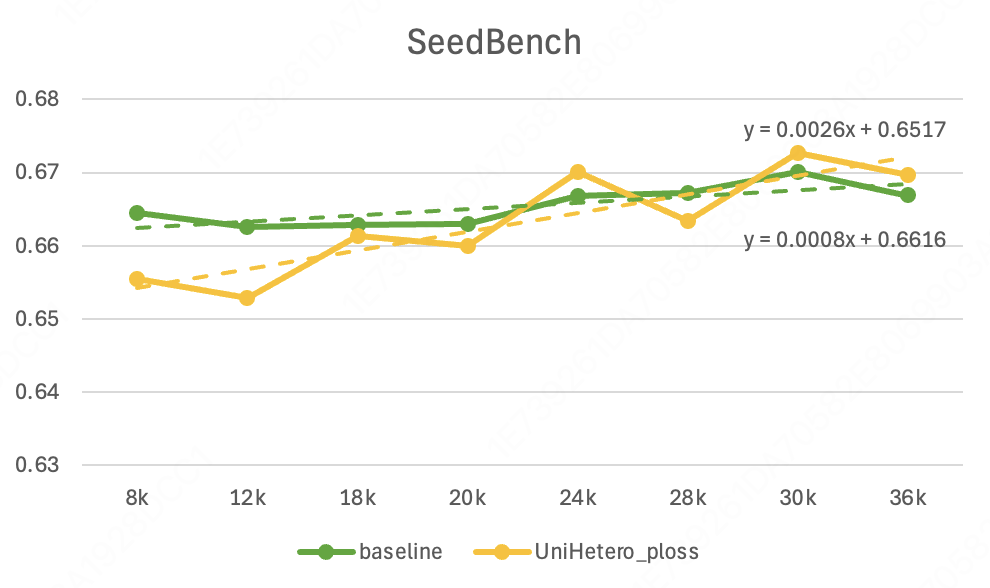}
    \end{subfigure}
    \caption{Performance of UniHetero with ploss using over 200M data samples. The x-axis represents the training steps. The dashed lines show the linear regression fit for the data scaling law.}
    \label{fig:data_scale}
\end{figure*}

The experimental results in Figure.~\ref{fig:data_scale} reveals that using only ploss (i.e. UniHetero\_ploss) shows a favorable scaling trend in understanding performances, eventually surpassing other model settings in later training stages, validating that generation enhances understanding.

\begin{figure*}[t]
    \centering
    \begin{subfigure}[b]{0.48\textwidth}
        \centering
        \includegraphics[width=\textwidth]{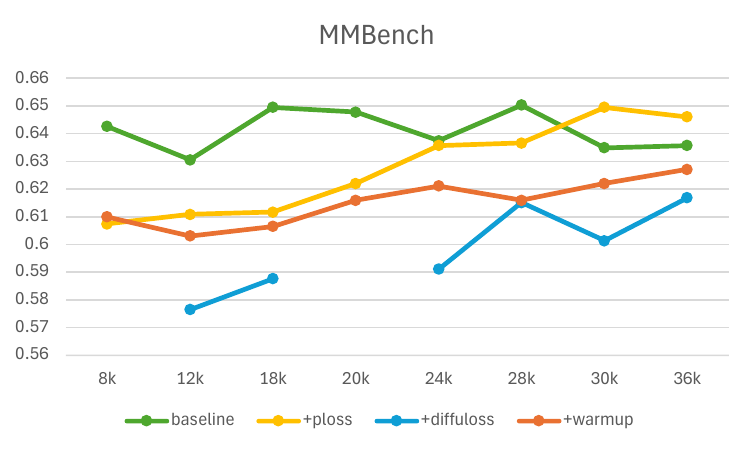}
    \end{subfigure} \hfill
    \begin{subfigure}[b]{0.48\textwidth}
        \centering
        \includegraphics[width=\textwidth]{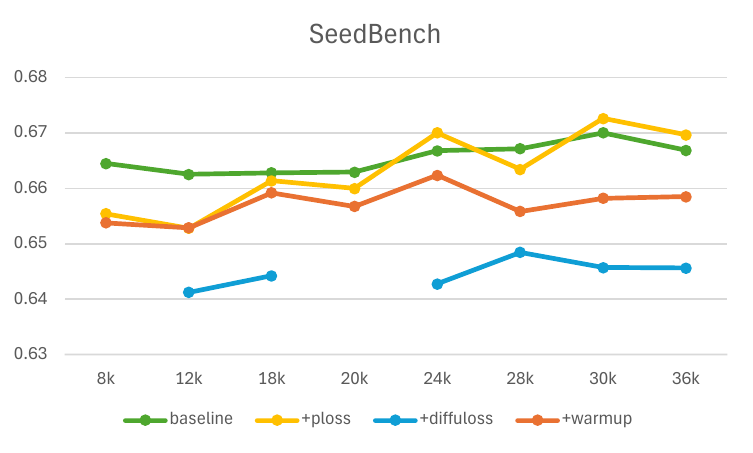}
    \end{subfigure}
\caption{Ablation results for different vision generation configurations. “+ploss” indicates UniHetero using only ploss. “+diffuloss” refers to UniHetero using both ploss and diffuloss. “+warmup” corresponds to UniHetero employing ploss, diffuloss, and a warmed-up diffusion module.}
    \label{fig:albation}
\end{figure*}

Furthermore, introducing diffuLoss causes a decline in understanding performance and weakens the scaling trend, as illustrated in Figure~\ref{fig:albation}. A conflict between semantic representation and pixel-level representation emerges, which is consistent with findings reported in previous studies~\citep{chen2025blip3o,wu2024janus}. It seems that the vision decoder module accomplishes the representation transformation through leveraging LLM rather than by itself. 

Could the transformation decrease the dependence on LLMs? To explore this further, an additional experiment (denoted "+warmup") was performed by increasing the size of the diffusion structure and applying a warm-up phase to the model parameters. In practice, we adopt the pretrained mar-base-f32 model~\citep{li2024mar}, which is concise than Stable Diffusion~\citep{esser2024stablediffu} and shows remarkable class-condition generation performance. The experimental findings show a substantial boost in performance on understanding tasks, suggesting that image generation can potentially be realized without relying on LLM-based compensation or intricate architectures.

\subsection{Module Selection Analysis}
\label{sec:module_selection}
The selection of the invert-projector architecture and loss function was rapidly validated through small-scale experiments. 

\begin{table}[!ht]
  \centering
  \resizebox{0.8\columnwidth}{!}{
     \begin{tabular}{l|c|c|c|c|c}
     \toprule
        Method & Loss & Invert-Projector & Target & MMBench & SeedBench \\
        \midrule
        diffu-mse & MSE Loss     & Diffusion(MAR-base)   & Vision Encoder & - & - \\
        mlp-mse   & MSE Loss     & 1x MLP          & Vision Encoder & - & - \\
        mlp-cos   & Cosine Loss  & 1x MLP          & Vision Encoder & 0.5833 & 0.6189 \\
        norm-3-mlp-cos & Cosine Loss & 3x MLP+norm (MetaMorph) & Vision Encoder & 0.5773 & 0.6502 \\
        ema-mlp-llm-cos & Cosine Loss & 1xMLP+momentum & Input Embedding & 0.6168 & 0.6456 \\
        \bottomrule
    \end{tabular}  
  }
  \caption{Different module settings on loss, target and the structure of invert-projector.}
  \label{tab:module_selection}
\end{table}

As shown in Table.~\ref{tab:module_selection}, we first compare mlp-cos, diffu-mse and mlp-mse on caption tasks for a quick observation. Results show that mlp-cos performs the best, which aligns with conclusions from existing studies~\citep{tong2024metamorph,wang2024ross,ge2025seedx}.




Next, we conduct finer experiments with end-to-end metrics (MMBench~\citep{liu2024mmbench} and SeedBench~\citep{li2023seedbench}) to select the module setting.
Experiments indicate that fitting the input to the LLM layer yields superior results compared to fitting the output of the Vision Encoder. Specifically, the ranking observed is: ema-mlp-llm-cos > norm-3-mlp-cos > mlp-cos. This confirms the hypothesis in Equation.~\ref{eq:auto_sematic_e}, namely that performing autoregression on the LLM input embeddings results in lower cumulative error and improved modality adaptation. To achieve a similar level of performance, the commonly adopted fitting vision encoder (Equation.~\ref{eq:auto_sematic_x}) requires a larger number of parameters under the norm-3-mlp-cos configuration.

\begin{figure}[t]
    \centering
    \includegraphics[width=0.8\linewidth]{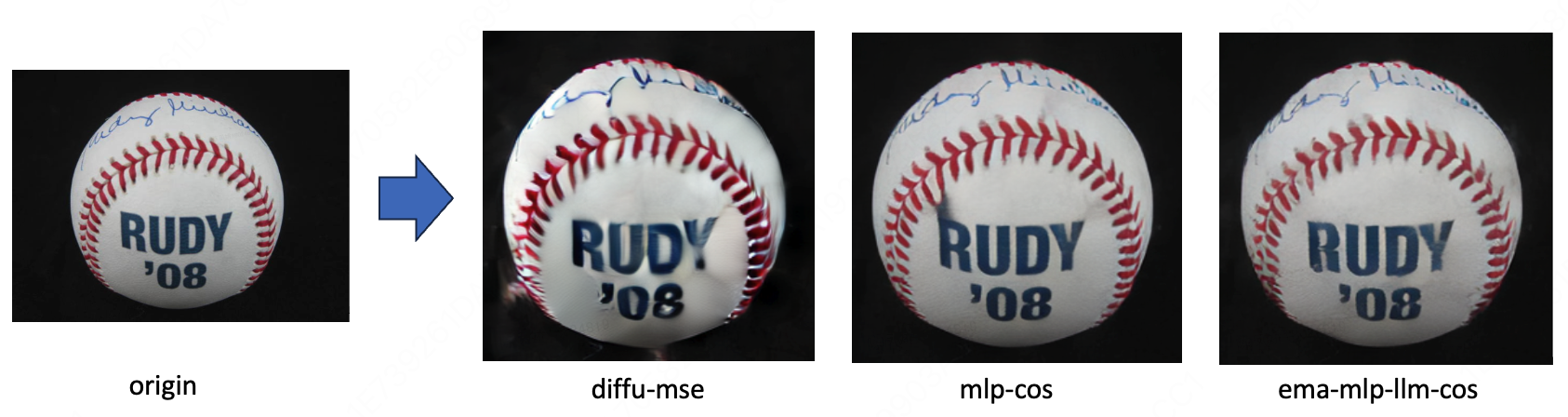}
    \caption{Visualization of overfitting generation on one image with various module setting (in Table.~\ref{tab:module_selection}).}
    \label{fig:gen_one_img}
\end{figure}

Furthermore, we also examine the effectiveness of pixel-level representations by overfitting on single image. The generation results, as shown in the Figure.\ref{fig:gen_one_img}, demonstrate that heterogeneous representations possess image generation capabilities and autoregression on input embedding is efficient to capture visual details. 

Finally, we mention that when fitting the LLM input, it is important to stabilize gradients via momentum-based updates or potentially by truncating gradient flow.

\subsection{Visualize Semantic Representation}
In the previous section's experiments, we observed that through autoregressive modeling of visual-semantic representations, generation can be verified to improve understanding on a large-scale dataset. Then, do visual-semantic representations actually capture? To investigate this, we now turn to a qualitative analysis.

Since visual-semantic representations cannot directly generate images, we adopt an indirect approach: first generate a visual representation from text, then use this representation to regenerate text, and examine whether the original textual information is preserved.

\begin{figure*}[t]
\centering
    \begin{subfigure}[b]{0.8\textwidth}
    \centering
    \includegraphics[width=\textwidth]{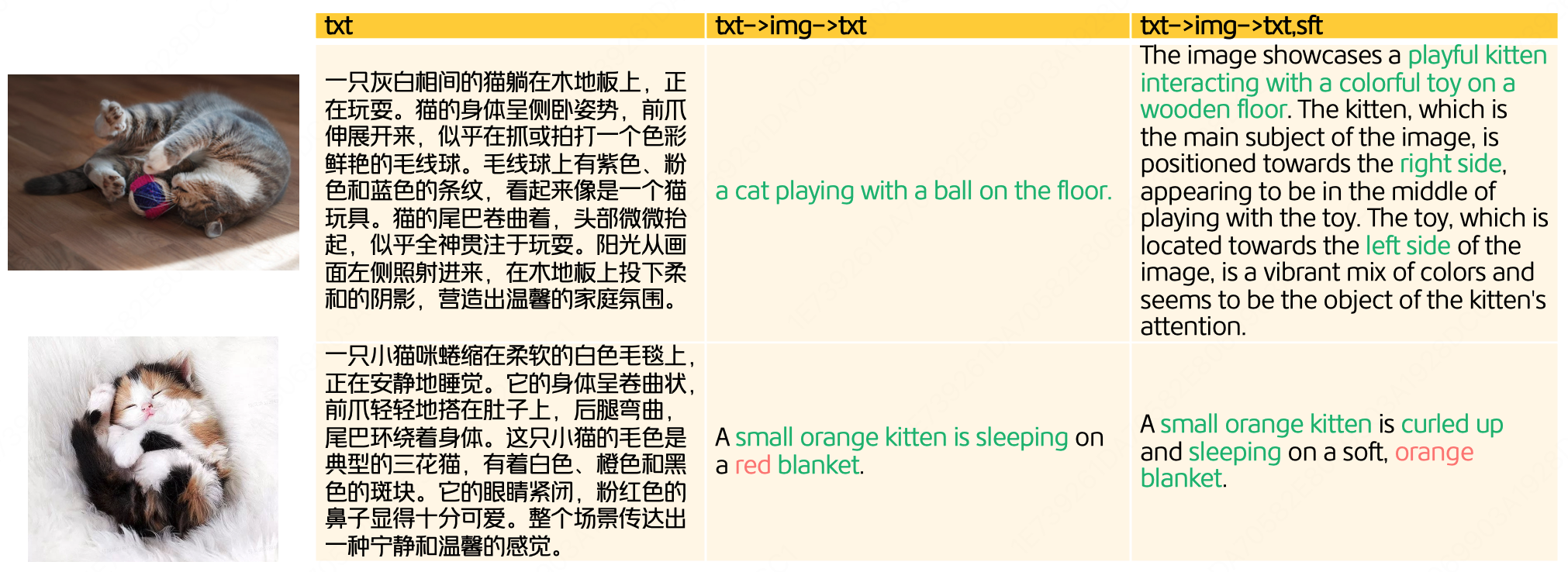}
    \caption{Visualize semantic representations on animals.}
    \label{fig:visual_case1}
    \end{subfigure}
\vspace{0.5em} 
    \begin{subfigure}[b]{0.8\textwidth}
    \centering
    \includegraphics[width=\textwidth]{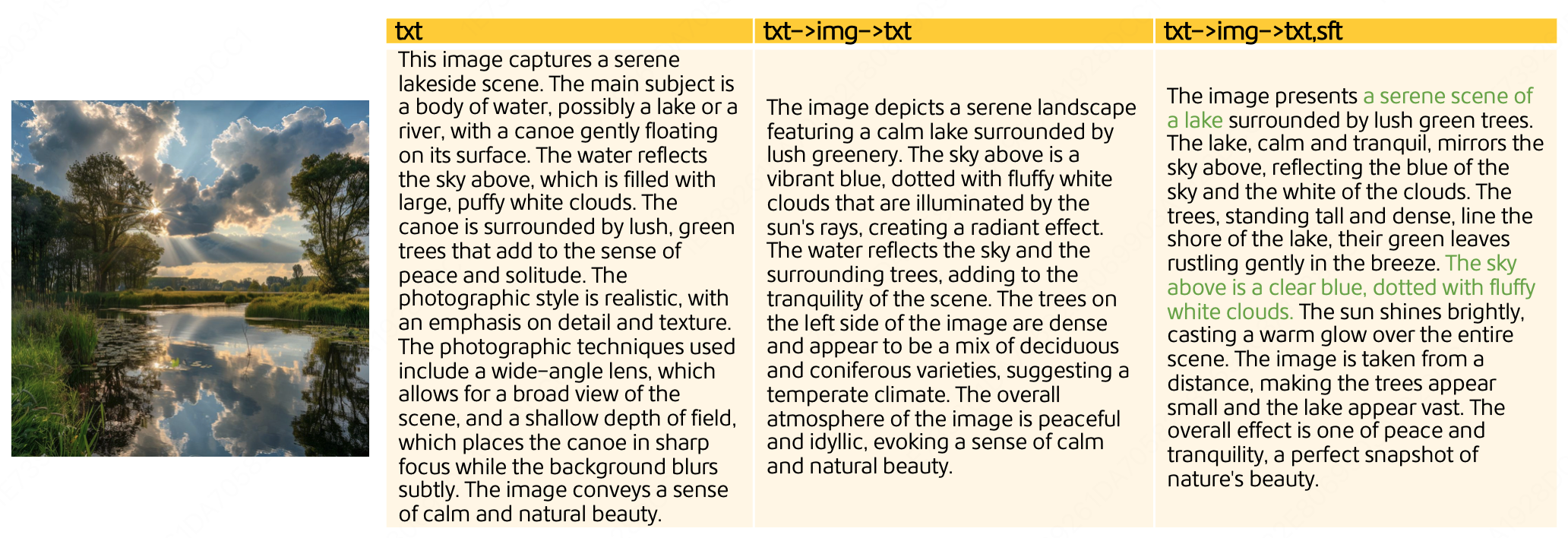}
    \caption{Visualize semantic representations on nature scenes.}
    \label{fig:visual_case2}
    \end{subfigure}
\vspace{0.5em} 
    \begin{subfigure}[b]{0.8\textwidth}
    \centering
    \includegraphics[width=\textwidth]{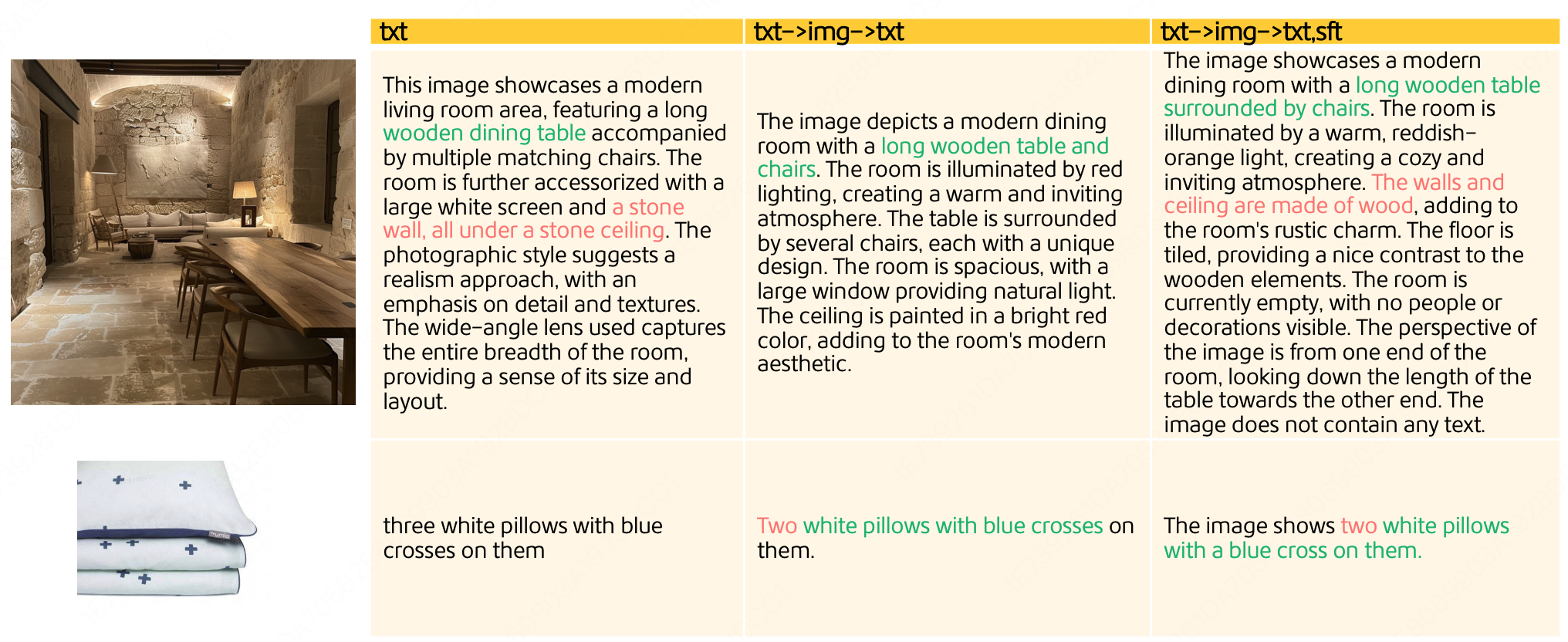}
    \caption{Visualize semantic representations badcases.}
    \label{fig:visual_case3}
    \end{subfigure}
\caption{Visualize semantic representations on different scenarios.}
\label{fig:visual_case_suite}
\end{figure*}

As shown in Figure.\ref{fig:visual_case_suite}, the first column displays the original images, the second column presents the captions of these images, and the third column shows the text regenerated from the visual representations generated based on these captions. Since this exploration was conducted directly after pretraining, the model's ability to follow instructions was not optimal. Therefore, a finetuned version is added in the fourth column. It can be observed that most image details, such as color, shape, and spatial position, are preserved through the semantic representations, though some recognition errors still exist.

\subsection{Visualize Pixel-Level Representation}
Although overfitting on single image works well in the Figure.\ref{fig:gen_one_img}, the quality of general image generation remains relatively low. While image generation quality is not the primary focus of this work, we have also explored potential directions for optimization.

We identify the discrepancy between training and inference as a major factor affecting generation performance. To mitigate this gap, we introduce an improved mask-rate scheduler for the training stage (Appendix~\ref{sec:gen_pretrain}) and an inference-time scaling strategy for the inference stage (Appendix~\ref{sec:gen_inference}). Both techniques yield performance gains, highlighting the promise of pixel-level image generation.



\section{Conclusion}
This work constructs UniHetero, a unified generation-understanding model under continuous encoding, which features a concise architecture and straightforward training. The study shows that the unified model attains a better data scaling pattern than models focused solely on image understanding, validating under large-scale pretraining that generation facilitates understanding. Preserving the LLM layer for semantic representation fusion may be key to learning multimodal knowledge.

The image semantic representations learn information such as objects, locations, shapes, and colors; the low-dimensional image representations enable pixel-level image generation. Image generation performance is further improved through pretraining data, masked sampling, and multi-round inference optimization, which showing potential for inference-time scaling.

\section*{Limitations}
In the future, we will continue our investigation along two main directions.

\textbf{Improve Pixel-level Generation.} While generation helps with semantic understanding, the model’s pixel-level visual synthesis remains weak. Potential improvements include using a larger vision decoder or employing a more effective warm-up schedule, adopting training strategies that mitigate negative side effects, and applying inference-time scaling techniques.

\textbf{Expand Semantic Representation Advantage.} The potential of learning from visual information remains insufficiently explored. Possible directions include acquiring physical knowledge from video data, generating semantic images as a form of visual CoT, and extracting visual knowledge without relying on aligned captions. Meanwhile, determining how to evaluate and verify what has been learned from such semantic representations is both essential and challenging.

\section*{Acknowledgement}

We hereby express our appreciation to the LongCat Team EVA Committee for their valuable assistance, guidance, and suggestions throughout the course of this work.

\bibliographystyle{unsrtnat}
\bibliography{custom}

\clearpage

\appendix

\section{Generation Improvement on Pretraining Stage}
\label{sec:gen_pretrain}
First, the current training data uses reversed image-understanding data as image-generation data without filtering for image quality or adjusting captions accordingly, which is not conducive to image generation quality. Therefore, by incorporating high-quality image generation data, we observed some improvement in image texture and details, but issues such as disharmonious distortions and blurring between tokens still persist.

\begin{figure*}
    \centering
    \includegraphics[width=\textwidth]{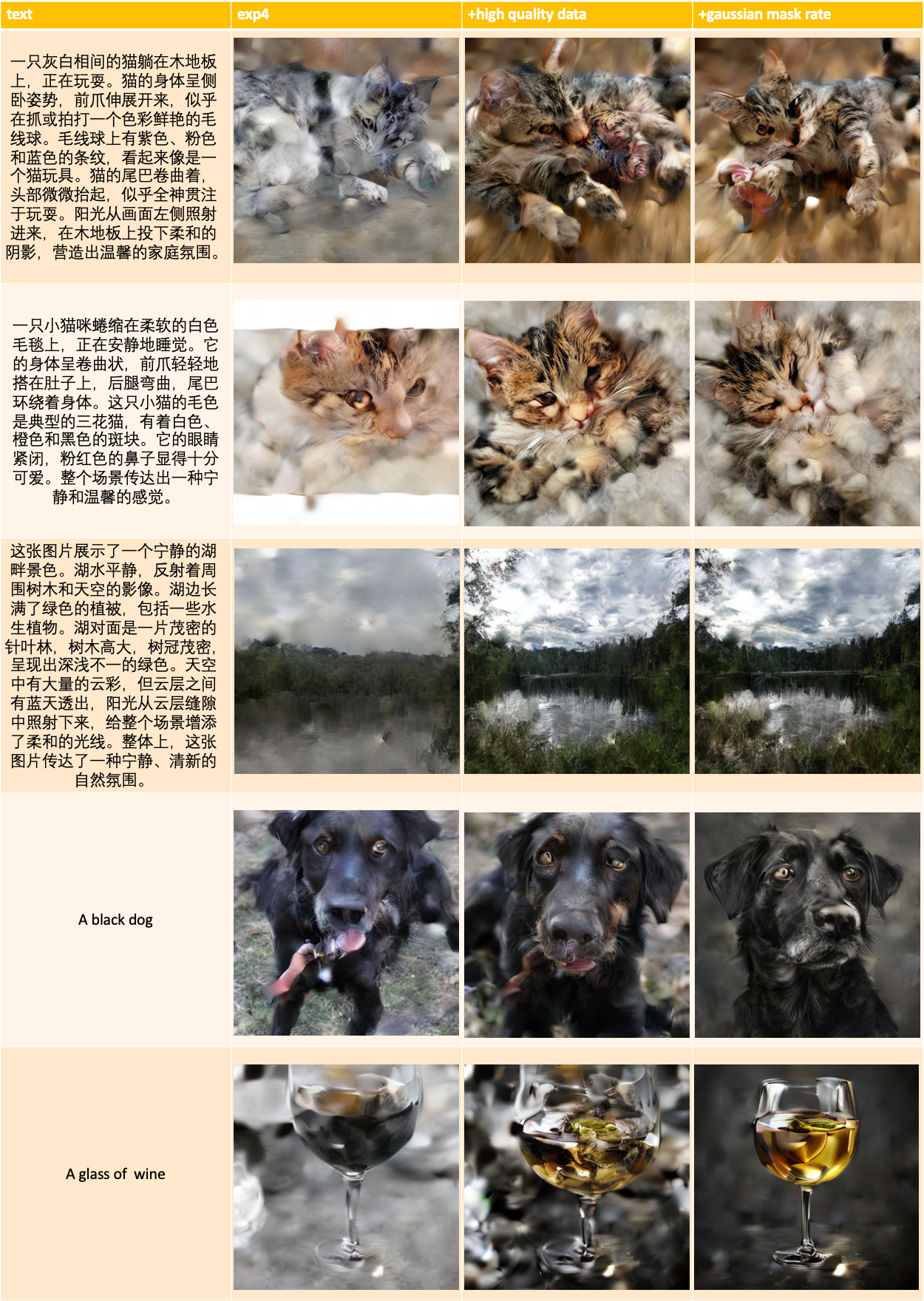}
    \caption{Visualize the generation improvement on pretraining stage.}
    \label{fig:gen_pretrain}
\end{figure*}

Considering the consistency between training and inference (the same as MAR inference), the masking probability has a significant impact. A higher masking probability can improve image generation learning but may negatively affect image understanding. Meanwhile, during inference, the masking probability decreases from 1.0 to 0, so the training process needs to expose the model to various masking probabilities. Therefore, we replaced the fixed masking probability in training with Gaussian sampling with a mean of 0.7. As shown in Figure.\ref{fig:gen_pretrain}, experimental results show a significant improvement in image generation quality.

\section{Generation Improvement on Inference Stage}
\label{sec:gen_inference}
During the inference stage, further consideration is given to the differences between training and inference. During training, it is assumed that unmasked tokens are correct and the average mask rate is around 70\%, but this assumption does not hold during inference, especially for the initial steps where prediction starts from 100\% masking, making the task more difficult and resulting in lower token quality, which significantly differs from the conditions in training. Therefore, we consider evaluating token generation quality after one round of generation, selectively retaining high-quality tokens, and regenerating low-quality ones until overall quality can no longer be improved. However, this requires training a token-level reward model. As an initial attempt, we randomly select tokens for regeneration, assuming that most generated tokens are of acceptable quality, and regenerate them either in their original order or with random masking. 

\begin{figure*}
    \centering
    \includegraphics[width=\textwidth]{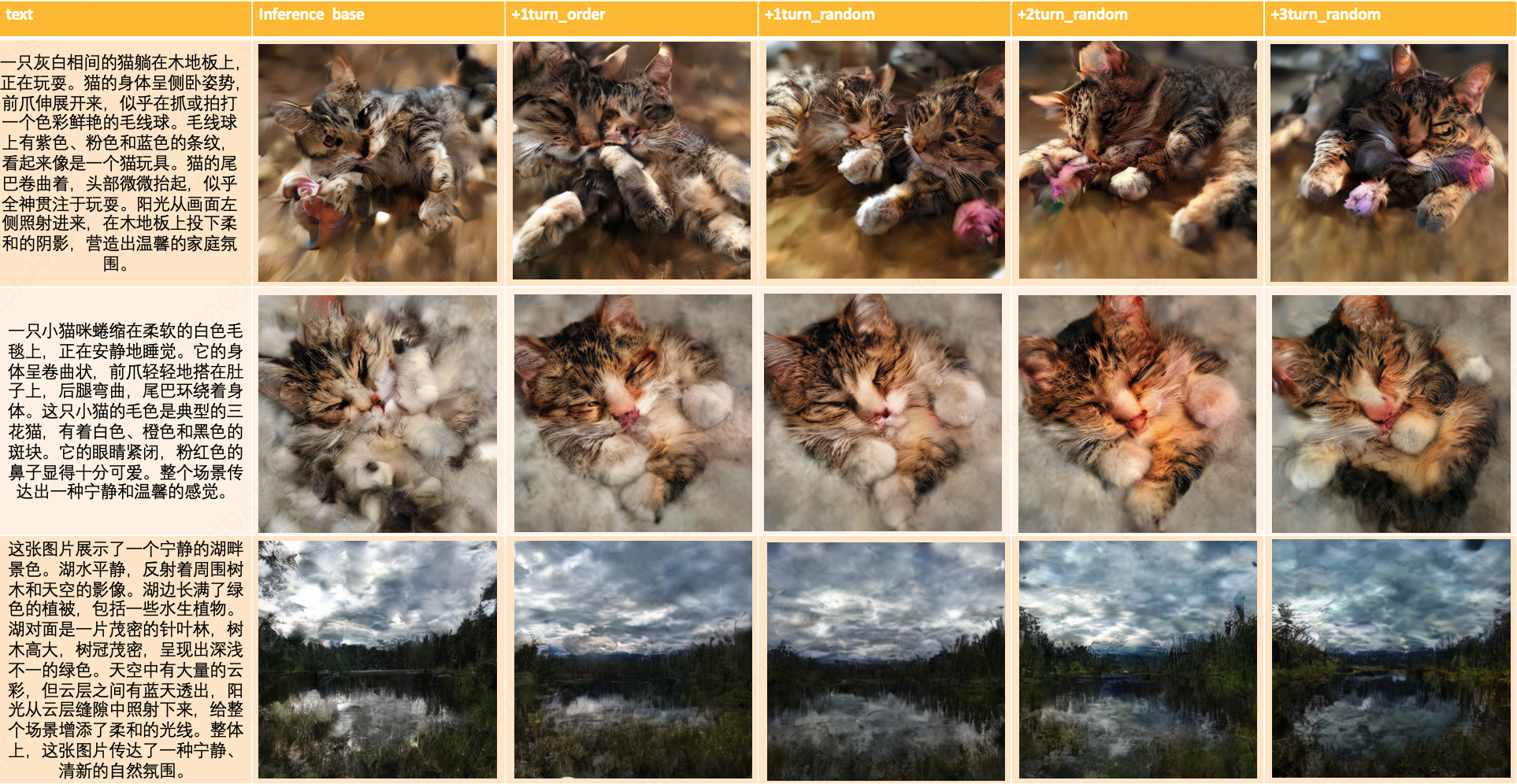}
    \caption{Visualize the generation improvement on inference stage. }
    \label{fig:gen_inference}
\end{figure*}

The experimental results are shown in the Figure.\ref{fig:gen_inference}. For most generated samples, quality improves after one random iteration, with issues such as distortion and blurry details being mitigated. However, there are cases where the original generated image was of good quality but degrades after regeneration, indicating that selective regeneration is still necessary. This aligns with the concept of "inference-time scaling" in the generation field~\citep{ma2025inferencetimescaling,guo2025generateimagescotlets}, suggesting that this direction has potential for further exploration.

\end{document}